\documentclass[lettersize,journal]{IEEEtran}
\usepackage{amsmath,amsfonts}
\usepackage{algorithmic}
\usepackage{algorithm}
\usepackage{array}
\usepackage[caption=false,font=normalsize,labelfont=sf,textfont=sf]{subfig}
\usepackage{textcomp}
\usepackage{stfloats}
\usepackage{url}
\usepackage{verbatim}
\usepackage{graphicx}
\usepackage{cite}
\usepackage{bm}
\usepackage{multirow}
\usepackage{xspace}
\usepackage{cleveref}
\usepackage{booktabs}
\usepackage{xcolor}
\usepackage{amssymb} 
\usepackage[T1]{fontenc}

\newcommand{\method}{DreamTalk\xspace}

\newcommand{\vect}[1]{\bm{#1}}

\newcommand{\eqword}[1]{{\text{#1}}}
\newcommand{\norm}[1]{\lVert #1 \rVert}

\usepackage{array}

\newcommand{\supp}{\textit{Supp. Mat.}\xspace}
\newcommand{\suppvideo}{\textit{Supp. Video}\xspace}

\newlength\savewidth\newcommand\shline{\noalign{\global\savewidth\arrayrulewidth
  \global\arrayrulewidth 1pt}\hline\noalign{\global\arrayrulewidth\savewidth}}

\begin{document}

\title{DreamTalk: When Emotional Talking Head Generation Meets Diffusion Probabilistic Models}

\author{Yifeng Ma,
        Shiwei Zhang,
        Jiayu Wang,
        Xiang Wang,
        Yingya Zhang,
        Zhidong Deng
\thanks{Yifeng Ma and Zhidong Deng are with Department of Computer Science and Technology, 
    BNRist, THUAI, State Key Laboratory of Intelligent Technology and Systems, Tsinghua University, Beijing 100084, China. (e-mail: mayf18@mails.tsinghua.edu.cn; michael@tsinghua.edu.cn).}
\thanks{Shiwei Zhang, Jiayu Wang and Yiyang Zhang are with Alibaba Group, Hangzhou 310023, China. (e-mail: \{zhangjin.zsw, wangjiayu.wjy, yingya.zyy\}@alibaba-inc.com).}
\thanks{Xiang Wang are with Huazhong University of Science and Technology, Wuhan 430074, China. (e-mail: wxiang@hust.edu.cn).}
\thanks{Yifeng Ma and Xiang Wang are interns at Alibaba Group.}
\thanks{Note: We would like to exclude the preprint titled "Dreamtalk: When expressive talking head generation meets diffusion probabilistic models"~\cite{ma2023dreamtalk} as prior art for the purpose of evaluating novelty, potential plagiarism, and self-plagiarism. This is because the preprint and the submitted manuscript are essentially the same article. We modified the preprint's title, added content, and then submitted it to the journal, but the core subject matter has not changed.}
}

\markboth{Journal of \LaTeX\ Class Files,~Vol.~14, No.~8, August~2021}%
{Shell \MakeLowercase{\textit{et al.}}: A Sample Article Using IEEEtran.cls for IEEE Journals}

\IEEEpubid{}

\maketitle

\begin{abstract}
Emotional talking head generation has attracted growing attention.
Previous methods, which are mainly GAN-based, still struggle to consistently produce satisfactory results across diverse emotions and cannot conveniently specify personalized emotions.
In this work, we leverage powerful diffusion models to address the issue and propose DreamTalk, a framework that employs meticulous design to unlock the potential of diffusion models in generating emotional talking heads.
Specifically, DreamTalk consists of three crucial components: a denoising network, a style-aware lip expert, and a style predictor.
The diffusion-based denoising network can consistently synthesize high-quality audio-driven face motions across diverse emotions.
To enhance lip-motion accuracy and emotional fullness, we introduce a style-aware lip expert that can guide lip-sync while preserving emotion intensity. 
To more conveniently specify personalized emotions, a diffusion-based style predictor is utilized to predict the personalized emotion directly from the audio, eliminating the need for extra emotion reference.
By this means, DreamTalk can consistently generate vivid talking faces across diverse emotions and conveniently specify personalized emotions.
Extensive experiments validate DreamTalk's effectiveness and superiority. The code is available at https://github.com/ali-vilab/dreamtalk.
\end{abstract}


\section{Introduction}
Audio-driven talking head generation, which concerns animating portraits with speech audio, has garnered significant interest due to its diverse applications, such as film dubbing, digital human generation, video conferences in band-limited conditions, and online education. 
To produce realistic talking heads, it is crucial to generate and control emotions. 
Recognizing that a single type of emotion can be expressed in diverse ways, recent research focus has shifted from modeling coarse-grained, discrete emotions to modeling more fine-grained, personalized emotions~\cite{wang2023progressive,ma2023styletalk,liang2022expressive}.
These personalized emotions are also called speaking styles~\cite{ma2023styletalk,wang2024styletalk++}. A speaking style is defined as facial motion patterns reflected in a talking video clip. In different video clips, the speakers may have different speaking habits and may display various emotions. Therefore, speaking styles are diverse.


Existing methods still struggle to 1) \emph{consistently} produce high-quality results across diverse speaking styles and 2) \emph{conveniently} specify desired speaking styles.
Existing methods~\cite{liang2022expressive,ji2022eamm,ma2023styletalk,gan2023efficient}
are mainly based on GANs~\cite{goodfellow2014generative}.
GANs' inherent issues, such as mode collapse and unstable training, impair their performance across diverse speaking styles.
Although these methods achieve satisfactory results for a limited range of speaking styles, they struggle with more diverse styles, especially ones unseen during training, often resulting in diminished emotional intensity, inaccurate lip motion, or sudden facial distortions~\cite{ma2023styletalk}.
Another issue is that to specify speaking styles, previous methods often rely on extra references, such as videos~\cite{liang2022expressive,ji2022eamm,wang2023progressive} or texts~\cite{ma2023talkclip,xu2023high,wang2024instructavatar}. 
Their acquisition requires extra manual effort and hence is inconvenient. 

\begin{figure}[t!]
  \centering
  \includegraphics[width=0.5\textwidth]{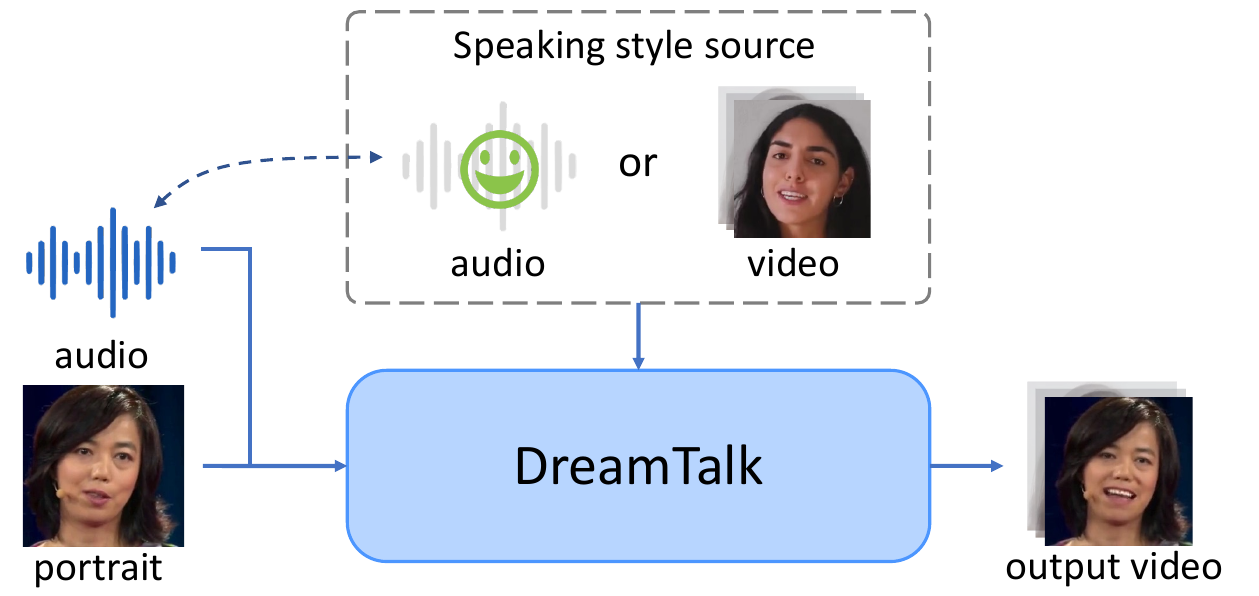}
  \caption{Leveraging the powerful diffusion models, \method can consistently generate high-quality talking heads across diverse speaking styles. Furthermore, \method can conveniently use audio to specify personalized speaking style, obviating the need for additional style references.}
  \label{fig:first_page}
\end{figure}

As a new line of generative technique, diffusion models~\cite{ho2020ddpm,sohl2015deep} have shown capability to produce high-quality results in numerous generative areas~\cite{dhariwal2021diffusion,rombach2022high,wang2023videocomposer,singer2022make,ao2023gesturediffuclip}.
The success of diffusion models, stemming from their superior properties such as powerful distribution learning~\cite{dhariwal2021diffusion,tevet2022humanmotiondiffusion}, make them exceptionally promising for exploring emotional talking head generation.
However, current diffusion-based talking head approaches~\cite{shen2023difftalk,yu2023talking,xu2024vasa,tian2024emo} primarily concentrate on generating talking heads with neutral expressions or a limited number of discrete emotions, lacking diverse and fine-grained speaking styles.
Therefore, exploring the full potential of diffusion models for generating talking heads with diverse speaking styles represents a promising, yet unexplored, research direction.


In this paper, we propose \method, an emotional talking head generation framework that takes advantage of diffusion models to consistently deliver high performance across diverse speaking styles and reduce the reliance on expensive style references. 
Specifically, \method is composed of a denoising network, a style-aware lip expert, and a style predictor. 
The diffusion-based denoising network produces audio-driven facial motions with the speaking style specified by a reference video.
The great distribution-learning property of diffusion models enable the denoising network to consistently produce high-quality results across diverse speaking styles.
To enhance the lip-sync, we design a style-aware lip expert that drives the denoising network to produce accurate lip motions under different speaking styles.
We observe that previous lip experts, which neglect emotional information, compromise the intensity of generated emotions. 
To preserve the emotion intensity, we find it important to integrate style information into the lip expert, thereby making it style-aware.
Finally, to eliminate the need for additional style references,
a diffusion-based style predictor is incorporated to predict personalized speaking styles directly from audio.
To predict more personalized emotions, we find it crucial to leverage the correlation between speaker identity and speaking styles; therefore, we provide identity information by incorporating the portrait as input.


The effectiveness of \method is demonstrated through comprehensive qualitative and quantitative evaluations. \method can even generate reasonable results for songs in multiple languages, despite these audios being significantly different from those in the training set. In summary, our contributions are as follows:

\begin{itemize}
    \item We propose \method, a diffusion-based framework that can consistently generate talking faces with precise lip-sync as well as rich emotions across diverse speaking styles. We find that diffusion models achieve better results than GANs for more diverse speaking styles.

    \item We explore how to use audio alone to predict personalized emotions, making it more convenient than relying on extra videos to specify speaking styles. We discover that incorporating identity information significantly enhances prediction accuracy.

    \item We propose a style-aware lip expert that can avoid reducing emotion intensity when providing lip guidance. We find that making the lip expert conditioning on speaking style information is crucial for maintaining emotional fullness. 

    \item Trained in a classifier-free manner, \method can use the classifier-free guidance scheme to adjust the intensity of arbitrary speaking styles. 
\end{itemize}


\section{Related Work}

\vspace{1mm}
\noindent\textbf{Audio-Driven Talking Head Generation.}
Audio-driven methods~\cite{das2020speech,guan2023stylesync,wang2023lipformer,sun2022masked,wang2023facecomposer} fall into two main categories: person-specific and person-agnostic. Person-specific approaches~\cite{suwajanakorn2017synthesizing,fried2019text,ji2021audio,wang2020mead,lu2021live} are constrained to generating videos for speakers seen during training. Many of these~\cite{yi2020audio, thies2020neural, song2020everybody,lah2021lipsync3d, ji2021audio, zhang20213d, zhang2021facial, tang2022memories} first craft 3D facial animations, later converting them into realistic videos. Recent advancements~\cite{guo2021ad,liu2022semantic,tang2022real,ye2023geneface,tang2022real,peng2024synctalk} have employed neural radiance fields for modeling, yielding high-fidelity, realistic videos. Conversely, person-agnostic methods~\cite{chung2017you,wang2022one,sadoughi2019speech,vougioukas2019realistic} target generating videos for unseen speakers. Early methods prioritized lip synchronization~\cite{song2018talking,chen2018lip,zhou2019talking,chen2019hierarchical,vougioukas2019realistic,prajwal2020lip}. Later works shifted focus to natural facial expressions~\cite{yu2023talking,zhou2020makelttalk} and head poses~\cite{chen2020talking,zhang2021flow,wang2021audio2head,zhou2021pose,zhang2023sadtalker}. FROND~\cite{sheng2023towards} introduces a fine-grained motion model that captures local facial movement keypoints and embeds overall motion context to predict audio-driven facial movements and achieve smooth temporal transitions. However, this method fails to generate emotional expressions during speech, thereby affecting the video's realism.

\vspace{1mm}
\noindent\textbf{Emotional Talking Head Generation.} 
Early methods~\cite{xu2023high,gan2023efficient,danvevcek2023emotional,gururani2023space,tan2023emmn,wang2020mead,ji2021audio,sinha2022emotion,peng2023emotalk} model expressions in discrete emotions.
To model fine-grained emotions, recent methods~\cite{ji2022eamm,liang2022expressive,ma2023styletalk,ma2023dreamtalk,tan2024edtalk} use an expression reference video and transfer the expressions from that video to the generated one. However, these GAN-based methods cannot consistently achieve high performance across diverse emotions. Our work addresses these issues by using diffusion models.

UniFaceGAN~\cite{cao2021unifacegan} introduces a temporally consistent facial video editing framework that handles both face swapping and face reenactment simultaneously by using a 3D reconstruction model and a novel temporal loss constraint.
Facial-Prior-Guided FME Generation~\cite{zhang2023facial} enhances facial micro-expression generation by utilizing adaptive weighted prior maps and facial priors to guide motion representation.
F3A-GAN~\cite{wu2021f3a} employs a 3D geometric flow, termed facial flow, to represent natural facial motion for continuous image synthesis.
Although all these methods are related to facial expression generation, none of them can generate accurate lip shapes driven by audio in different emotional contexts.

Conveniently specifying desired speaking styles is also important. Most previous methods rely on reference videos~\cite{ji2022eamm,liang2022expressive,ma2023styletalk} or text~\cite{ma2023talkclip,xu2023high,gan2023efficient}, which needs human labor. A more user-friendly approach is to derive speaking styles from the input audio. Previous methods can only infer a limited number of discrete emotion classes from audio ~\cite{ji2021audio,xu2023high,sinha2022emotion}. TH-PAD~\cite{yu2023talking} generates expressions only aligned with the audio rhythm, not aligning with the emotional content of the audio. Besides, previous methods neglect information in the input portrait. In this work, we aim to infer personalized emotions using input audio and portraits.

\vspace{1mm}
\noindent\textbf{Diffusion Models.}
Diffusion models~\cite{sohl2015deep,ho2020ddpm} have demonstrated strong performance across multiple vision tasks~\cite{dhariwal2021diffusion,wang2023videocomposer,zhang2023i2vgen,gao2024multi,wang2024headdiff,welker2024driftrec}, including text-to-image generation~\cite{ruiz2022dreambooth}, human motion generation~\cite{tevet2022humanmotiondiffusion}, and video generation~\cite{singer2022make,qing2023hierarchical,wei2023dreamvideo,wang2023videolcm,wang2023modelscope}. 
Most diffusion-based methods for talking head generation~\cite{shen2023difftalk,stypulkowski2023diffused,bigioi2023speech,mukhopadhyay2023diff2lip,xu2023multimodal,du2023dae,yu2023talking,wei2024aniportrait,wang2024v,liu2024anitalker,chen2024echomimic,he2023gaia}, including EMO~\cite{tian2024emo}, AniPortrait~\cite{wei2024aniportrait} and Hallo~\cite{xu2024hallo}, mainly generate talking heads with neutral emotions and lacks emotional controllability. Besides, the inference of EMO is slow. VASA~\cite{xu2024vasa} can only generate a limited number of emotions, lacking diverse, fine-grained speaking styles. In this work, we aim to harness diffusion models for generating and controlling diverse, fine-grained speaking styles in talking heads, presenting a more intricate challenge.

\section{Method}
\begin{figure*}[t!]
    \centering
    \includegraphics[width=1.0\linewidth]{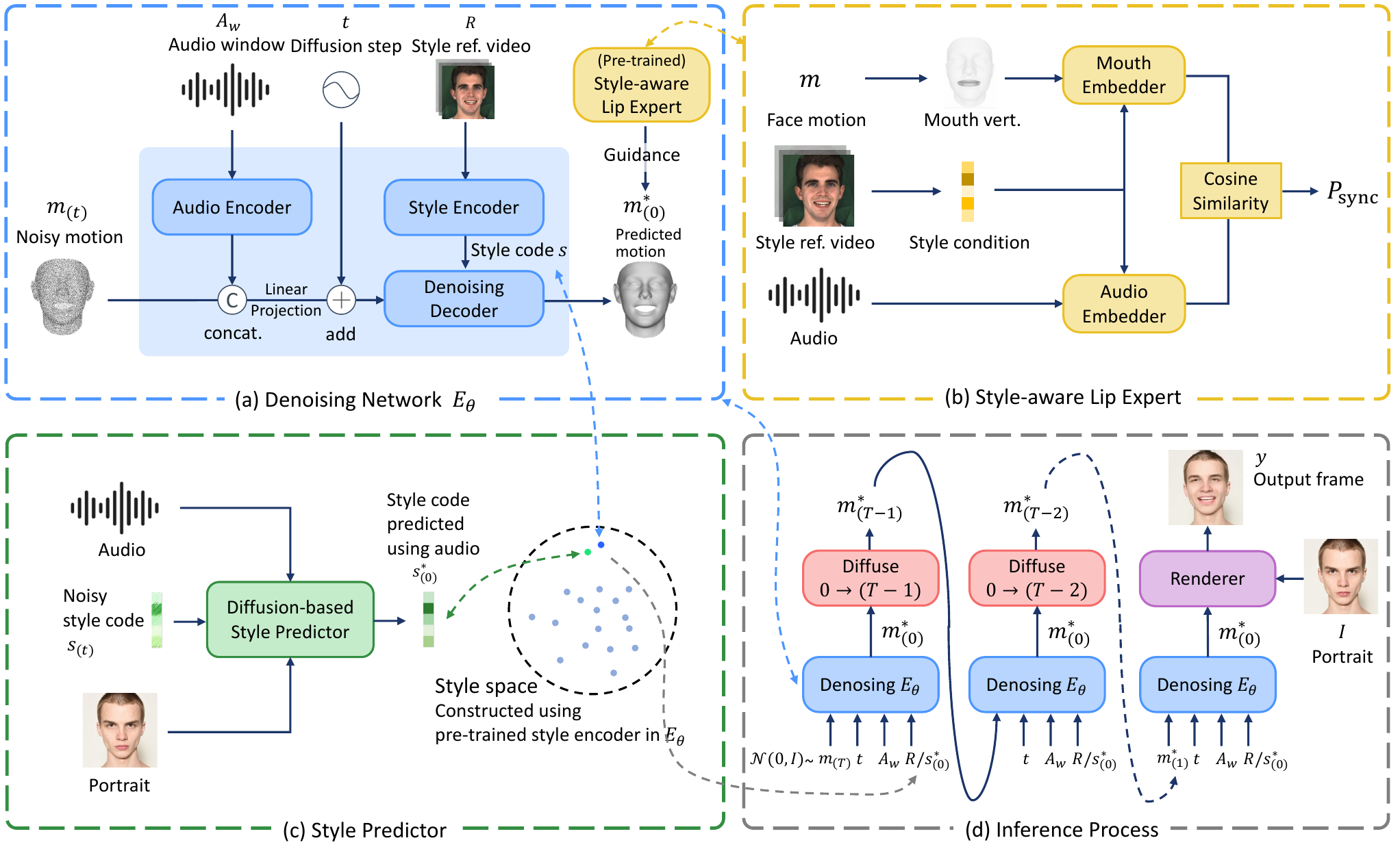}
    \caption{ 
    Illustration of DreamTalk. 
    A style-aware lip expert (b) is first trained to provide lip motion guidance for the denoising network (a). 
    The denoising network is then trained to predict emotional audio-driven face motions. Then,  A style predictor (c) is trained to use audio to predict the style code. During inference (d), the speaking style can be specified using style codes that are extracted from videos or derived from audio.
    }
    \label{fig:dreamtalk_pipeline}
\end{figure*}

\subsection{Problem Formulation}

Given a portrait $\vect{I}$, a speech $\vect{A}$, and a style reference video $\vect{R}$, our method aims to generate a talking head video with lip motions synchronized with the speech and the speaking style reflected in the reference video. The audio $\vect{A}=[\vect{a}_i]_{i=1}^{L}$ is parameterized as a sequence of acoustic features. $\vect{R}$ is a sequence of video frames. The head motions in the generated videos can originate from real videos or be produced by existing methods~\cite{zhang2023sadtalker,wang2024styletalk++}.

Besides, to conveniently specify speaking styles, our method also aims to infer the speaking style using solely the speech and the portrait, obviating the need for extra style references. 
The inferred speaking style can replace the role of style reference videos in controlling expressions (\cref{fig:first_page}), enabling our method to generate personalized emotions with only speech and portrait.

\subsection{DreamTalk}

As illustrated in \cref{fig:dreamtalk_pipeline}, \method comprises 3 key components: a denoising network, a style-aware lip expert, and a style predictor.

The denoising network computes face motion conditioned on the speech and style reference video. The face motion $\vect{M}=[\vect{m}_l]_{l=1}^{L}$ is parameterized as a sequence of expression parameters from 3D Morphable Models~\cite{blanz1999morphable}. The face motion is rendered into video frames by a renderer~\cite{ren2021pirenderer}. The style-aware lip expert provides lip motion guidance under diverse expressions and thus drives the denoising network to achieve accurate lip-sync while preserving emotion fullness. The style predictor can predict the speaking style aligned with that conveyed in speech.

\vspace{1mm}
\noindent\textbf{Denoising Network.} The denoising network synthesizes face motion sequence frame-by-frame in a sliding window manner. It predicts  a motion frame $\vect{m}_l$ using an audio window $\vect{A}_w = [\vect{a}_i]_{i=l-w}^{l+w}$, where $w$ denotes the window size.

The denoising network leverages forward and reverse diffusion processes. The diffusion process is modeled as a Markov noising process. Starting from a motion frame $\vect{m}_{(0)}$, it incrementally introduces Gaussian noise into the real data, gradually diffusing towards a distribution resembling $\mathcal{N}(\vect{0}, \vect{I})$. Consequently, the distribution evolves as follows:
\begin{equation}
    q(\vect{m}_{(t)} | \vect{m}_{(t-1)}) = \mathcal{N}(\sqrt{\alpha_n}\vect{m}_{(t-1)}, (1-\alpha_n)\vect{I}),
\end{equation}
where $\vect{m}_{(t)}$ is the motion frame sampled at diffusion step $t$, $t\in\{1, \dots, T\}$, and $\alpha_n$ is determined by the variance schedules. Conversely, the reverse diffusion process, or the denoising process, predicts the added noise in a noisy motion frame. Starting from a random motion frame $\vect{m}_{(T)} \sim \mathcal{N}(\vect{0}, \vect{I})$, the denoising process incrementally removes the noise and recovers the original motion $\vect{m}_{(0)}$.

Instead of predicting the noise as formulated by~\cite{ho2020ddpm}, we follow \cite{ramesh2022hierarchical} and predict the signal itself. The denoising network ${E}_{\theta}$ predicts $\vect{m}_{(0)}$ based on the noisy motion, the diffusion step, the speech context, and the style reference:
\begin{equation}
\label{eqn:denoising_network}
    \vect{m}^{*}_{(0)} = E_{\theta}(\vect{m}_{(t)}, t, \vect{A}_w, \vect{R}).
\end{equation}
The asterisk($*$) indicates quantities that are generated.

Our denoising network has a transformer architecture~\cite{vaswani2017transformer}. The audio window $\vect{A}_w$ is first fed into a transformer-based audio encoder and the output is concatenated with the noisy motion $\vect{m}_{(t)}$ in the channel dimension. After linearly projected to the same dimension, the concatenated results and the timestep $t$ are summed and served as the key and value of a transformer decoder. To extract the speaking style from the style reference,  a style encoder first extracts the sequence of 3DMM expression parameters from $\vect{R}$ and then feeds them into a transformer encoder. The output tokens are aggregated using a self-attention pooling layer~\cite{safari2020self} to obtain the style code $\vect{s}$. The style code is repeated $2w+1$ times and added with positional encodings. The results serve as the query of the transformer decoder. The middle output token of the decoder is fed into a feed-forward network to predict the signal $\vect{m}_{(0)}$.

\vspace{1mm}
\noindent\textbf{Style-aware Lip Expert.} We observe that using solely the denoising loss in standard diffusion models results in inaccurate lip motions. We conjecture that the loss alone is insufficient for the denoising network to effectively focus on generating precise lip motions. A typical remedy is to involve a pre-trained lip expert~\cite{prajwal2020lip} that provides lip motion guidance. However, we observe the lip expert reduces the intensity of expressions. This stems from the fact that the lip expert merely focuses on a generic speaking style, which leads to generating face motions in a uniform style. 

To address this issue, we introduce a style-aware lip expert. The proposed lip expert is trained to evaluate lip-sync under diverse speaking styles. Therefore, it can provide lip motion guidance under diverse speaking styles and strike a better balance between style expressiveness and lip-sync. The lip expert $\mathcal{E}$ computes the probability that a clip of audio and lip motions are synchronous conditioned on style reference $\vect{R}$:
\begin{equation}
    P_{\eqword{sync}} = \mathcal{E}([\vect{a}_i]_{i=l}^{l+n}, [\vect{m}_i]_{i=l}^{l+n}, \vect{R}),
\end{equation}
where $n$ denotes the clip length.


The style-aware lip expert encodes the lip motions and audio into respective embeddings conditioned on style reference and then computes the cosine similarity to represent the sync probability. To obtain lip motion information from face motion $\vect{m}$, we first convert $\vect{m}$ into the corresponding face mesh and select vertices in the mouth area as the representation of the lip motion. The lip motion and audio encoders are mainly implemented by MLPs and 1D-convolutions, respectively. The style condition is fused into embeddings by first extracting style features from style reference using a style encoder, which mirrors the architecture of the one in the denoising network but does not share parameters with it, and then concatenating the style features with intermediate feature maps from embedding encoders.

\vspace{1mm}
\noindent\textbf{Style Predictor.} Specifically, the style predictor $S_\phi$ predicts the style code $\vect{s}$ extracted by the style encoder in the trained denoising network. Since we observe that style codes correlate with speaker identity (\cref{sec:style_code_visualization}), the style predictor also integrates the portrait as input. The style predictor is instantiated as a diffusion model and is trained to predict the style code itself:
\begin{equation}
    \vect{s}^*_{(0)} = S_\phi(\vect{s}_{(t)}, t, \vect{A}, \vect{I}),
\end{equation}
where $\vect{s}_{(t)}$ is the style code sampled at diffusion step $t$.


The style predictor $S_\phi$ is a transformer encoder on a sequence consisting of, in order: audio embeddings, an embedding for the diffusion timestep, a speaker info embedding,  the noised style code embedding, and a final embedding called learned query whose output is used to predict the unnoised style code. Audio embeddings are audio features extracted using self-supervised pre-trained speech models. To obtain the speaker info embedding, our method first extracts the 3DMM identity parameters, which include the face shape information but removes irrelevant information, such as expressions, from the portrait, and then embeds it into a token using an MLP.

\vspace{1mm}
\noindent\textbf{Discussion: Advantages over StyleTalk.} Although StyleTalk, a GAN-based baseline, also leverages transformer modules, DreamTalk presents notable advantages: 1) StyleTalk's modules and loss functions are overly complex, which may cause unstable generation results, while DreamTalk's are simple, making it more extensible and robust. Since GAN's mode-collapse issue hampers modeling diverse speaking styles, to enhance emotion intensity, StyleTalk uses a complex dynamic network and up to six loss terms. As discussed in \cref{sec:main_results}, the overly complex design may cause unstable and inferior results. In contrast, DreamTalk, leveraging the power of diffusion models, does not need complex modules and only uses two loss terms, which is much simpler. 2)  StyleTalk makes incorrect assumptions about the data, which may impair the performance. To apply losses that enhance emotion intensity, StyleTalk assumes the speaking styles are consistent in a predefined video group. However, as discussed in \cref{sec:style_code_visualization}, these speaking styles are actually varied. DreamTalk does not need such an assumption. 3) StyleTalk can only specify speaking styles using videos, while DreamTalk, leveraging the style predictor, can specify styles only using input audio, which is more convenient.

\begin{table*}[t]
    \centering
    \caption{Quantitative Comparisons. We do not receive GC-AVT samples on Voxceleb2.  SA is only evaluated on MEAD for emotional methods.}
    \resizebox{\textwidth}{!}{%
    \renewcommand{\arraystretch}{1.2}
    \begin{tabular}{c|cccccccc}
    \toprule
    \multirow{2}{*}{Methods} & \multicolumn{5}{c}{MEAD / HDTF / Voxceleb2}   \\
        & SSIM$\uparrow$ & CPBD$\uparrow$ & F-LMD$\downarrow$ & M-LMD$\downarrow$ & $\eqword{Sync}_{\eqword{conf}}$ $\uparrow$  & SA$\uparrow$  \\ 
        \shline
        
        MakeItTalk~\cite{zhou2020makelttalk} &        0.73/0.57/0.52 & 0.11/0.24/0.24 & 3.97/5.12/6.29 & 5.32/4.55/5.15 & 2.10/3.16/2.17 & -  \\ 
        Wav2Lip~\cite{prajwal2020lip} &        0.80/0.63/0.54 & \textbf{0.18}/0.30/\textbf{0.30} & 2.72/4.53/5.85 & 4.05/3.60/4.64 & \textbf{5.26}/\textbf{5.83}/\textbf{5.70} & -  \\ 
        PC-AVS~\cite{zhou2021pose} &        0.50/0.42/0.36 & 0.07/0.13/0.09 & 5.83/9.71/12.9 & 4.97/4.17/7.42 & 2.18/4.85/4.73 & -  \\ 
        AVCT~\cite{wang2022one} &        0.83/0.76/0.64 & 0.14/0.22/0.23 & 2.92/2.86/3.62 & 5.52/3.57/3.71 & 2.53/4.27/3.89 & -  \\ 
        GC-AVT~\cite{liang2022expressive} &        0.34/0.36/\ \  - \,\, & 0.14/0.28/\ \  - \,\, & 8.04/10.2/\ \  - \,\, & 7.10/6.23/\ \  - \,\, & 2.42/4.72/\ \  - \,\, & 18.7 \\ 
        EAMM~\cite{ji2022eamm} &        0.40/0.40/0.43 & 0.08/0.14/0.20 & 6.70/7.03/6.36 & 6.48/6.86/4.89 & 1.41/2.54/2.24 & 20.1  \\ 
        StyleTalk~\cite{ma2023styletalk} &        0.84/0.81/0.66 & 0.16/0.30/0.29 & 2.12/1.96/2.92 & 3.25/2.41/2.96 & 3.47/4.82/4.51 & 74.2  \\ 
        SadTalker~\cite{zhang2023sadtalker} &        0.69/0.77/0.44 & 0.16/0.24/0.19 & 4.12/5.99/9.12 & 4.37/4.07/6.11 & 2.76/4.35/4.38 & -  \\ 
        PD-FGC~\cite{wang2023progressive} &        0.49/0.41/0.35 & 0.05/0.13/0.12 & 5.50/9.50/12.5 & 4.10/4.23/8.19 & 2.27/4.68/4.64 & 43.9  \\ 
        EAT~\cite{gan2023efficient} &        0.53/0.59/0.47 & 0.15/0.26/0.20 & 5.54/3.86/5.53 & 4.79/4.03/5.88 & 2.16/4.54/4.35 & 36.4  \\ 
        \hline
        \textbf{DreamTalk} &        \textbf{0.86}/\textbf{0.85}/\textbf{0.69} & 0.16/\textbf{0.31}/\textbf{0.30} & \textbf{1.93}/\textbf{1.80}/\textbf{2.69} & \textbf{2.91}/\textbf{2.15}/\textbf{2.72} & 3.78/5.17/4.90 & \textbf{86.7} \\ 
        \textcolor{gray}{Ground Truth} &        \textcolor{gray}{1/1/1} & \textcolor{gray}{0.22/0.31/0.33} & \textcolor{gray}{0/0/0} & \textcolor{gray}{0/0/0} & \textcolor{gray}{4.13/5.44/5.23} & \textcolor{gray}{92.5}  \\ 
    \bottomrule
    \end{tabular}
        }%

    \label{tab:quanitative}
\end{table*}

\subsection{Training and Inference} 
\label{sec:dreamtalk_training}

\noindent\textbf{Training.} The style-aware lip expert is first pre-trained by determining whether randomly sampled audio and lip motion clips are synchronous as in~\cite{prajwal2020lip} and then frozen during training the denoising network. We use cosine-similarity with binary cross-entropy loss during training. Specifically, we compute cosine-similarity for the face motion embedding $\vect{e}^m$ and audio embedding $\vect{e}^a$ to represent the probability that the input audio-motion pair is synchronized. The training loss of the lip expert is: 
\begin{equation}
    \mathcal{L}_{\eqword{expert}} =\eqword{BCE}( \frac{\vect{e}^m\cdot \vect{e}^a}{\max (||\vect{e}^m||_2\cdot||\vect{e}^a||_2 )}).
\end{equation}

The denoising network  $E_{\theta}$ is trained by sampling random tuples $(\vect{m}_{(0)},t,\vect{A}_w,\vect{R})$ from dataset, corrupting $\vect{m}_{(0)}$ into $\vect{m}_{(t)}$ by adding Gaussian noises, executing denoising steps to $\vect{m}_{(t)}$, and optimizing the loss:
\begin{equation}
    \mathcal{L}_{\eqword{net}} = 
    \lambda_{\eqword{denoise}}\mathcal{L}_{\eqword{denoise}} +
    \lambda_{\eqword{sync}}\mathcal{L}_{\eqword{sync}}.
\end{equation}
Specifically, the ground-truth motion $\vect{m}_{(0)}$, and the speech audio window $\vect{A}_w$ are extracted from the training video of the same moment. $t$ is drawn from the uniform distribution $\mathcal{U}\{1,T\}$. The style reference $\vect{R}$ is a video clip randomly drawn from the same video containing $\vect{m}_{(0)}$.

We first compute the denoising loss of the diffusion models~\cite{ho2020ddpm} defined as:
\begin{equation}
    \mathcal{L}_{\eqword{denoise}} = \norm{\vect{m}_{(0)} - E_{\theta}(\vect{m}_{(t)}, t, \vect{A}_w, \vect{R})}_{2}^{2}.
\end{equation}
Then, the denoising network maximizes the synchronous probability via a sync loss on generated clips:
\begin{equation}
    \mathcal{L}_{\eqword{sync}} = -\eqword{log}(P_{\eqword{sync}}).
\end{equation}

Classifier-free guidance~\cite{ho2022classifier} is used to train our model. Specifically, $E_\theta$ is trained to learn both the style-conditional and unconditional distributions via randomly setting $\vect{R} = \varnothing$  by $10\%$ chance during training. $\varnothing$ is implemented as a sequence of face motions $[\vect{m}_i]$ with all zero values. For inference, the predicted signal is computed by
\begin{equation}
\begin{aligned}
    \vect{m}^{*}_{(0)} & =  \omega E_{\theta}(\vect{m}_{(t)}, t, \vect{A}_w, \vect{R}) \\
    & + (1 - \omega) E_{\theta}(\vect{m}_{(t)}, t, \vect{A}_w, \varnothing),
\end{aligned}
\end{equation}
instead of \cref{eqn:denoising_network}. This approach enables controlling the effectiveness of the style reference $\vect{R}$ through adjustment of the scale factor~$\omega$.


When training the style predictor, we draw a random video, then extract audio $\vect{A}$ and style code $\vect{s}_{(0)}$ (using the trained style encoder) from it. Since 3DMM identity parameters may leak expression information, the portrait $\vect{I}$ is sampled from another video with the same speaker identity. The style predictor $E_\phi$ is trained by optimizing the loss:
\begin{equation}
    \mathcal{L}_{\eqword{pred}} = \norm{\vect{s}_{(0)} - S_{\phi}(\vect{s}_{(t)}, t, \vect{A}, \vect{I})}_{2}^{2},
\end{equation}

We utilize PIRenderer \cite{ren2021pirenderer} as the renderer and fine-tune it on emotional dataset to enable it to generate emotional talking faces.

\vspace{1mm}
\noindent\textbf{Inference.}
Our method can specify speaking styles using either reference videos or solely through input audio and portrait.
In the case of reference videos, style codes are derived using the style encoder in the denoising network. When relying solely on input audio and portrait, the style predictor takes these inputs and employs a denoising procedure to obtain the style code. 

With the style code, the denoising network utilizes the sampling algorithm of DDPM~\cite{ho2020ddpm} to produce face motions. It first samples a random motion $\vect{m}_{(T)}^*\sim{}\mathcal{N}(\vect{0}, \vect{I})$ then computes denoised sequences $\{\vect{m}_{(t)}^*\},{t=T-1,\dots,0}$ by incrementally removing the noise from $\vect{m}_{(t)}^*$. Finally, the motion $\vect{m}_{(0)}^*$ is the generated face motion. The sampling process can be accelerated by leveraging DDIM~\cite{song2020DDIM}. The output face motions are then rendered into videos by the renderer.



\section{Experiments}

\begin{figure*}[t]
  \centering
  \includegraphics[width=0.98\linewidth]{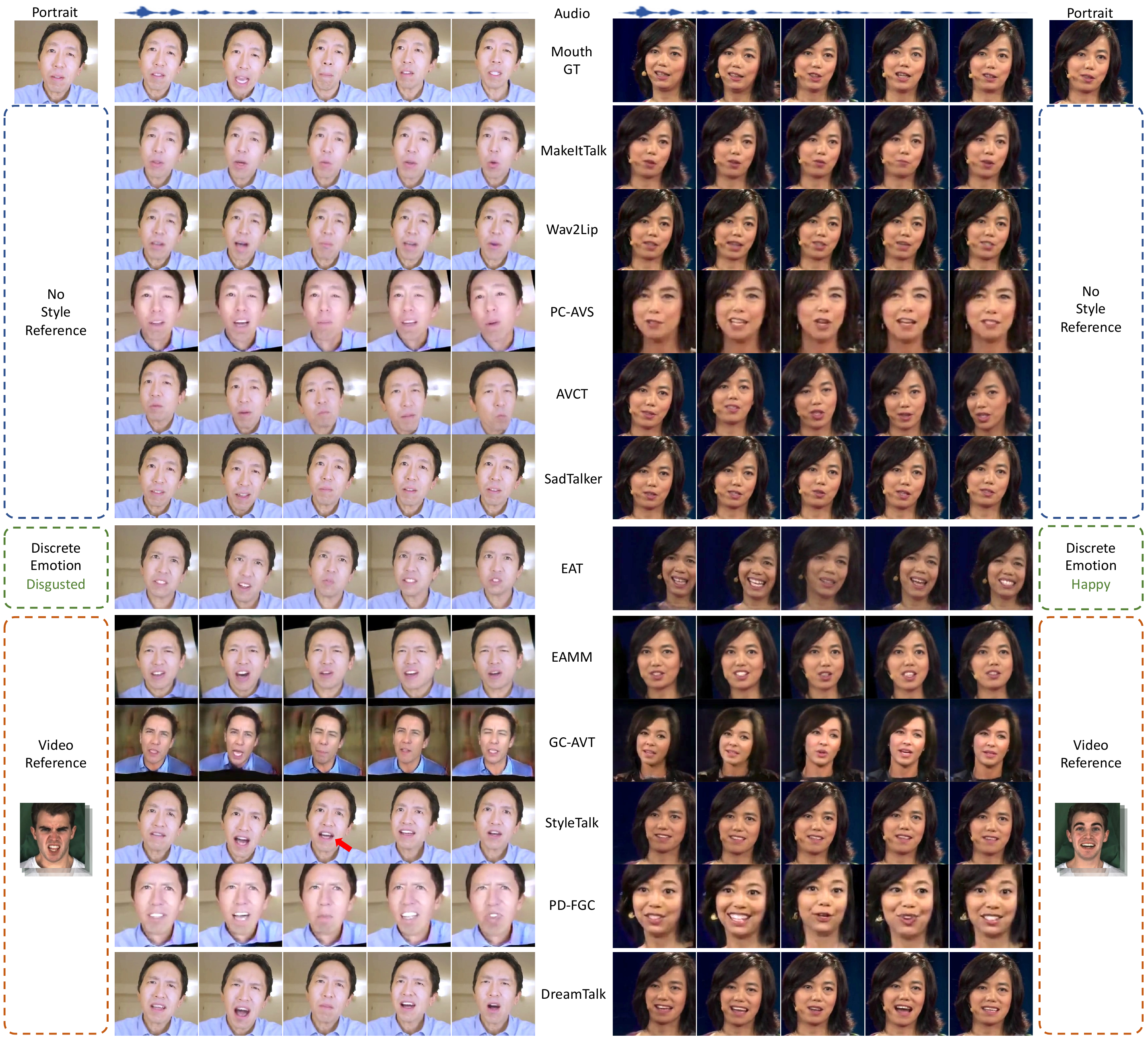}
  \caption{Qualitative comparisons. The \emph{red arrow} indicates inaccurate lip motions.}
  \label{fig:qualitative}
\end{figure*}

\subsection{Experimental Setup}
\label{sec:experimental_setup}
\noindent \textbf{Datasets.}
We train and evaluate the denoising network on MEAD~\cite{wang2020mead}, HDTF~\cite{zhang2021flow}, and Voxceleb2~\cite{chung2018voxceleb2}.  Since  Voxceleb2 official videos are of low resolution, we redownload the original YouTube videos and re-crop the videos. The style-aware lip expert is trained on MEAD and HDTF. We train the style predictor on MEAD and evaluate it on MEAD and RAVEDESS~\cite{livingstone2018ryerson}. 

\vspace{1mm}
\noindent \textbf{Baselines.}
We compare our method with previous methods: MakeitTalk~\cite{zhou2020makelttalk}, Wav2Lip~\cite{prajwal2020lip}, PC-AVS~\cite{zhou2021pose}, AVCT~\cite{wang2022one}, GC-AVT~\cite{liang2022expressive},  EAMM~\cite{ji2022eamm},  StyleTalk~\cite{ma2023styletalk}, DiffTalk~\cite{shen2023difftalk},  SadTalker~\cite{zhang2023sadtalker}, PD-FGC~\cite{wang2023progressive}, and EAT~\cite{gan2023efficient}. DiffTalk's released model cannot generate reasonable results until submission, so we perform qualitative comparisons using videos from its demo. For other methods, we generate the samples using released models or with authors' help. When generating samples, we use the audio and the first image from the test video as inputs. We use a segment of the test video as the reference. Except when evaluating the style predictor, the style of DreamTalk is specified by the video. 

\vspace{1mm}
\noindent\textbf{Metrics.} To evaluate video quality, we use SSIM~\cite{wang2004image} and the CPBD~\cite{narvekar2009no}. To evaluate lip-motion accuracy, we use the SyncNet confidence score ($\text{Sync}_{\text{conf}}$)~\cite{chung2016out} and the Landmark Distance around mouth area (M-LMD)~\cite{chen2019hierarchical}. To evaluate the accuracy of generated expressions, we use the Landmark Distance on the full face (F-LMD) and a newly proposed metric Style Accuracy (SA). SA is the accuracy obtained from classifying samples using a speaking style classifier. When training the classifier, we divide the MEAD dataset into several groups with approximately consistent speaking styles and train the classifier to sort videos into the correct groups. Therefore, if a method generates accurate expressions, its samples will be classified into correct group and hence it will get higher SA. The details of SA metric are reported in \supp.



\subsection{Main Results}
\label{sec:main_results}

\noindent\textbf{Quanitative Comparisons.} 
As shown in \cref{tab:quanitative}, our method outperforms previous methods across most metrics. 
Wav2Lip's SyncNet confidence score is higher than ours, even surpassing the ground truth. This is because Wav2Lip is trained using SyncNet as a discriminator.
Notably, our method's SyncNet confidence score closely aligns with the ground truth, and it achieves the best M-LMD scores, which indicates its capability for precise lip synchronization. Furthermore, our superior performance in the F-LMD and SA metrics demonstrates our method's proficiency in generating facial expressions consistent with the reference speaking style.

\noindent\textbf{Qualitative comparisons.}
\cref{fig:qualitative} shows the qualitative comparisons. The portraits, style references, and audio are all unseen during training. 

\begin{figure*}[t]
    \centering\small
    \begin{minipage}[t!]{0.48\textwidth}
        \centering
        \includegraphics[width=\textwidth]{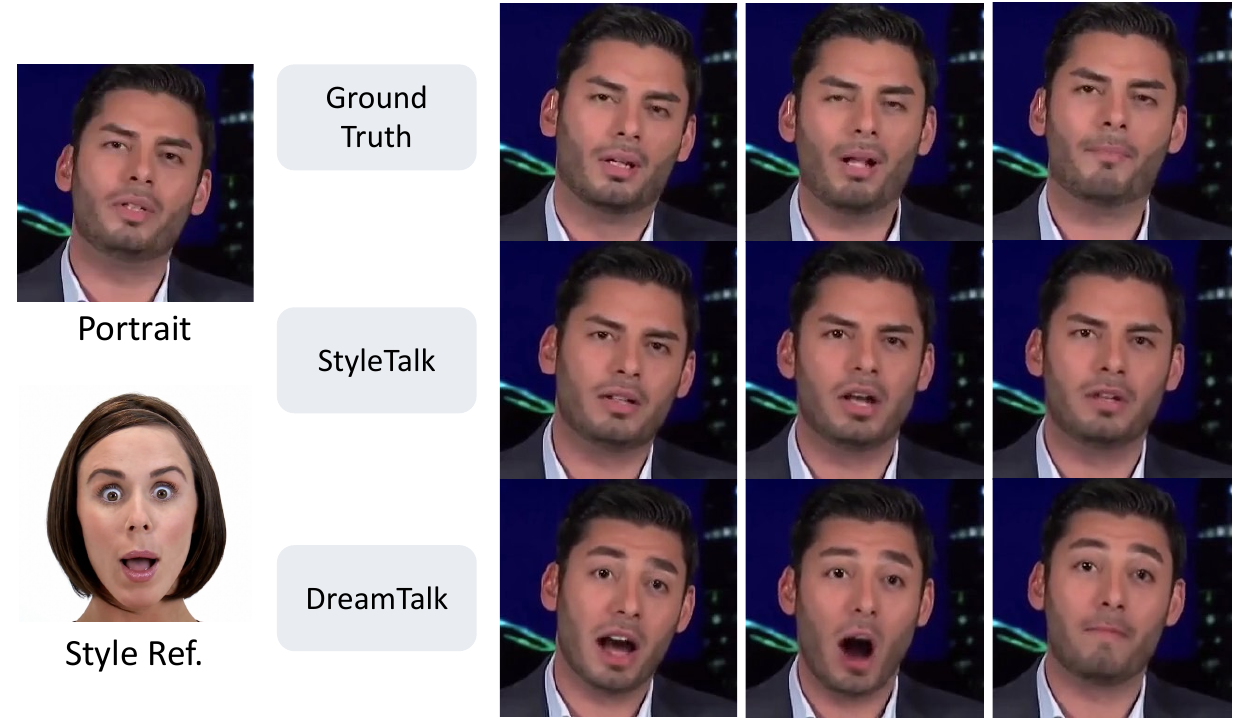}
        \caption{Comparisons with StyleTalk on in-the-wild style reference. StyleTalk fails to generate accurate emotion.}
        \label{fig:styletalk_in_the_wild}
    \end{minipage}
    \hskip1em
    \begin{minipage}[t!]{0.48\textwidth}
        \centering
        \includegraphics[width=\textwidth]{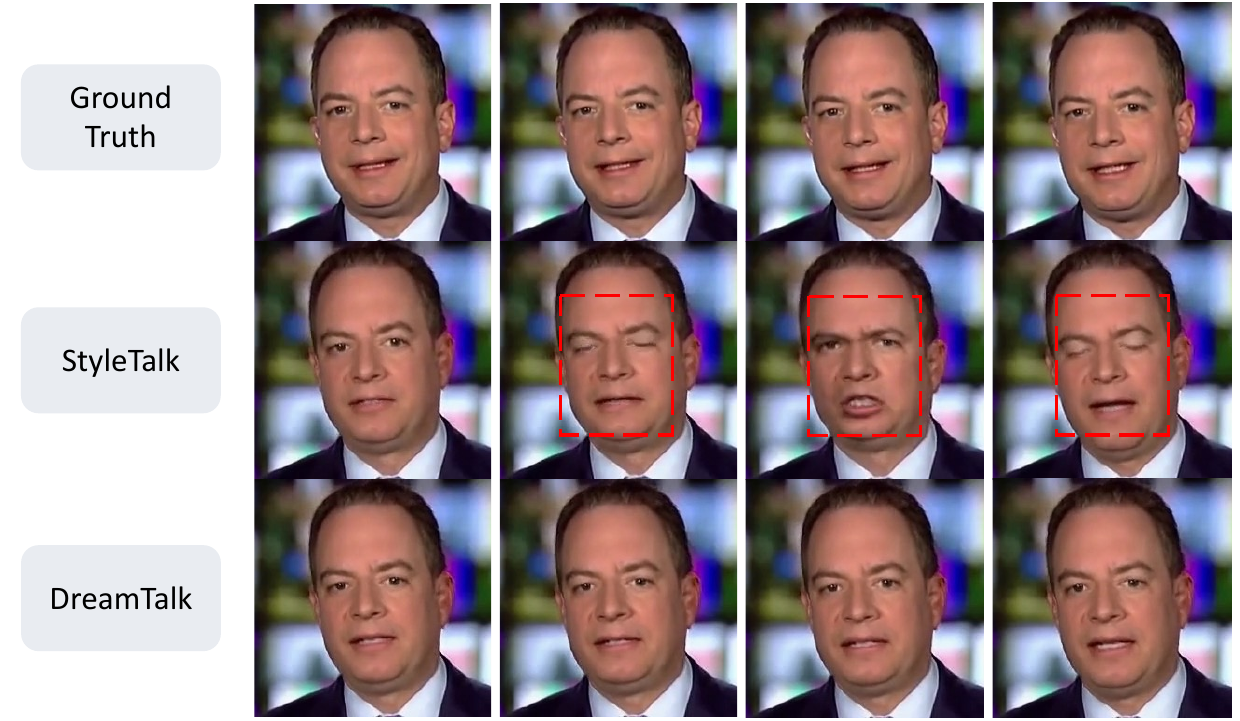}
        \caption{Sudden face distortion (\emph{marked in red box}) that frequently occurs in StyleTalk's output. Better viewed in \suppvideo}
        \label{fig:styletalk_face_distortion}
    \end{minipage}
\end{figure*}

\begin{table*}[t]
\centering
\caption{Comparisons of identity preservation on MEAD. DreamTalk's score is competitive to non-emotional methods and is the best in emotional methods.}
\resizebox{\textwidth}{!}{%
\begin{tabular}{cccccccccccccc}
\toprule  
Method &  MakeitTalk  & Wav2Lip & PC-AVS & AVCT & SadTalker & GCAVT & EAMM & StyleTalk & PD-FGC & EAT & TH-PAD  & Ours & Ground Truth \\
\midrule
CSIM $\uparrow$ & 0.77 & 0.89 & 0.17 & 0.79 & 0.78 & 0.20 & 0.36 & 0.64 & 0.35 & 0.61 & 0.30 & 0.71 & 0.80 \\

\bottomrule 
\end{tabular}
}

\label{table:CSIM}
\end{table*}

StyleTalk, one of the most competitive baselines, fails to consistently generate high-quality results across diverse speaking styles. Firstly, StyleTalk frequently generates videos with sudden facial distortions (\cref{fig:styletalk_face_distortion}), making it unsuitable for large-scale practical applications. We do not observe such phenomena in \method's results. We speculate that the reason for Styletalk's unstable generation is that StyleTalk uses overly complex loss functions and is based on GANs, models that suffer from unstable training. Secondly, StyleTalk fails to generate rich emotions, especially for in-the-wild style references. As shown in \cref{fig:qualitative}, the output from StyleTalk exhibits discrepancies with the style reference: the left speaker's eyes are not as narrowed, and the right speaker's mouth is not opened as widely. The discrepancies are more pronounced when using in-the-wild style references. As shown in \cref{fig:styletalk_in_the_wild}, Styletalk fails to generate expressions consistent with those in the style reference, including raised eyebrows, glaring eyes, and widened mouths. Initially, we speculate that these issues are caused by the limited training data of StyleTalk, but we observe that when trained only using the same data as StyleTalk, DreamTalk can still generate expressions consistent with the style reference (the result is also shown in \cref{fig:styletalk_in_the_wild}). Therefore, we speculate that the reason is that GAN's mode-collapse issue impairs the performance across diverse speaking styles. Thirdly, StyleTalk's lip-sync is inferior. A notable example is shown in \cref{fig:qualitative}: when the speaker utters "m"; the expected closed-mouth motion is replaced by an open mouth in StyleTalk's output.

\begin{figure*}[t]
    \centering\small
    \begin{minipage}[t!]{0.48\textwidth}
        \centering\small
        \includegraphics[width=\columnwidth]{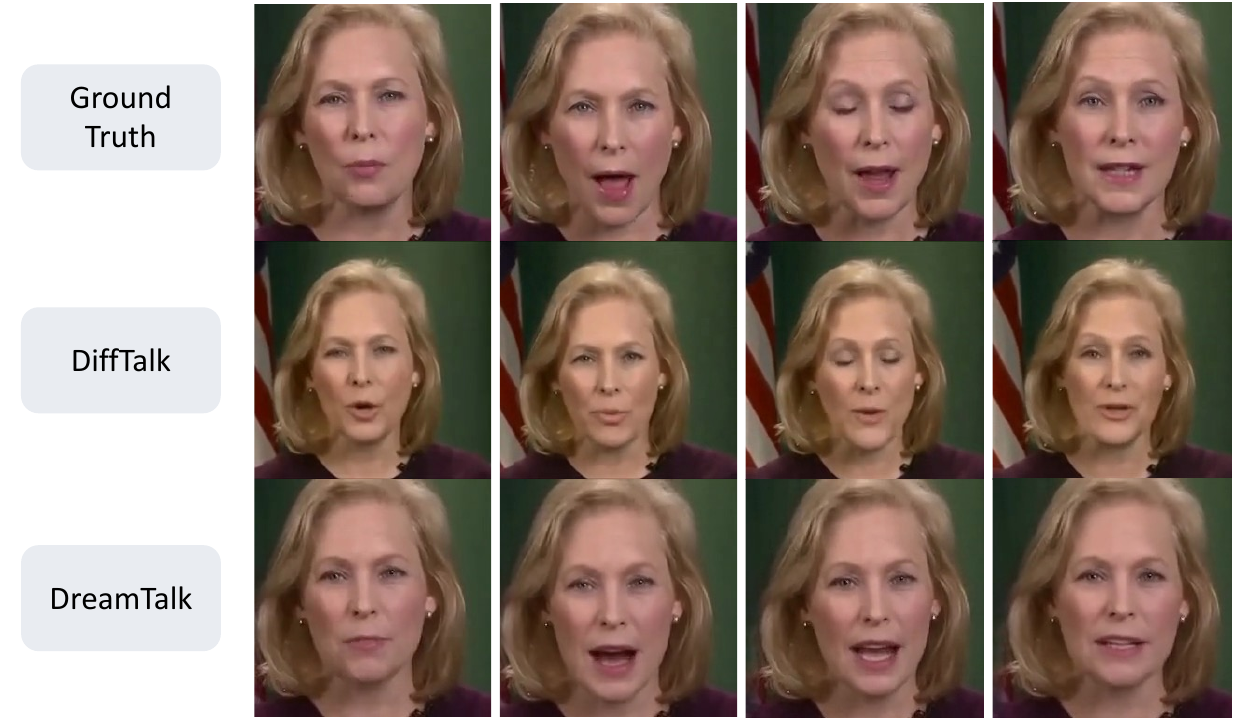}
        \caption{%
        Comparisons with DiffTalk. DiffTalk fails to achieve lip-sync and produces jitteriness.
        }
        \label{fig:compare_with_difftalk}
    \end{minipage}
    \hskip1em
    \begin{minipage}[t!]{0.48\textwidth}
        \centering\small
        \includegraphics[width=\columnwidth]{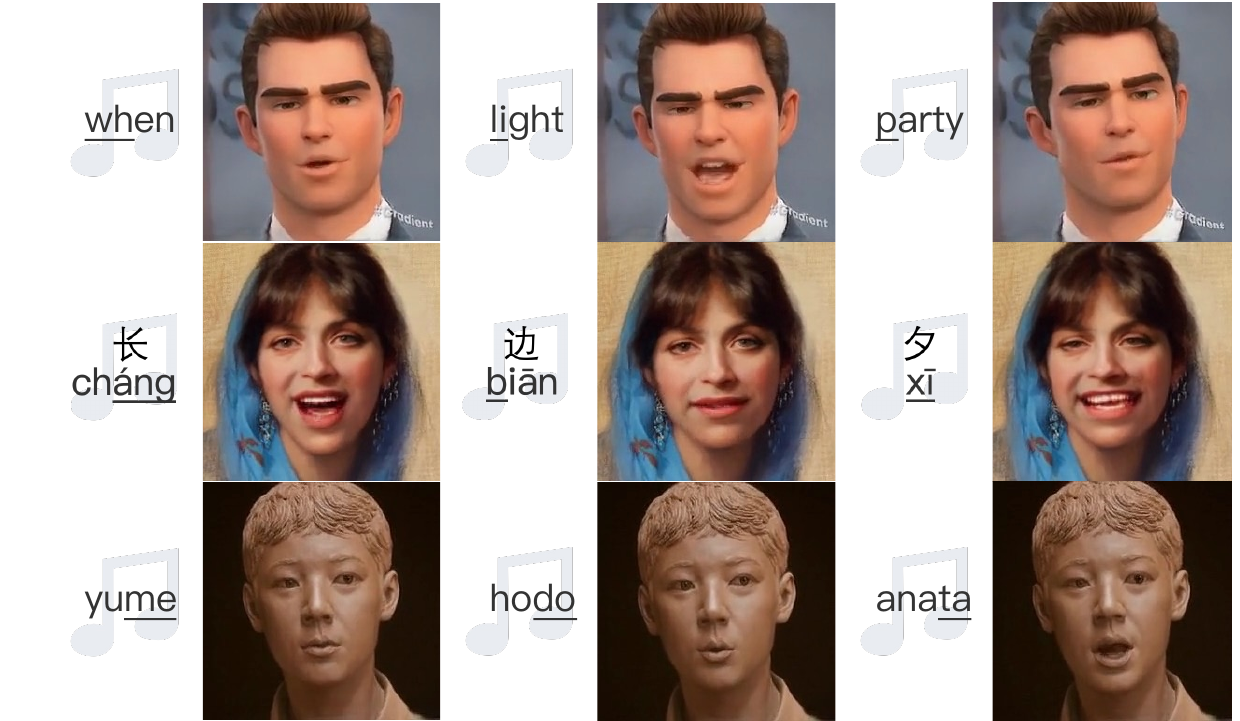}
        \caption{%
        Emotional results generated from songs in multiple languages (English, Chinese, Japanese).
        }
        \label{fig:songs}
    \end{minipage}
\end{figure*}

It can be seen that MakeItTalk and AVCT struggle with accurate lip synchronization. While Wav2Lip and PC-AVS synchronize lips accurately, their outputs appear blurry. SadTalker, on the other hand, generally aligns lip movements with audio but occasionally displays unnatural jitters.
EAT can only generate discrete emotions, lacking the finesse for nuanced expressions. For example, in the left case, the style reference shows the speaker narrowing his eyes, but EAT merely produces a generic disgusted look with wide-open glaring eyes. Additionally, as shown in the right case, EAT struggles to maintain a consistent face shape during speaker head movements.
EAMM, GC-AVT, PD-FGC can produce fine-grained emotions. However, EAMM falls short in lip synchronization, GC-AVT and PD-FGC struggle with preserving speaker identity, and all three have issues rendering a plausible background. 
As shown in \cref{fig:compare_with_difftalk}, DiffTalk struggles with lip synchronization and produces jitteriness and artifacts in the mouth area.

In contrast, \method excels in producing realistic talking faces that not only mirror the reference speaking style but also achieve precise lip synchronization and superior video quality.



\begin{figure*}[t!]
  \centering
  \includegraphics[width=\textwidth]{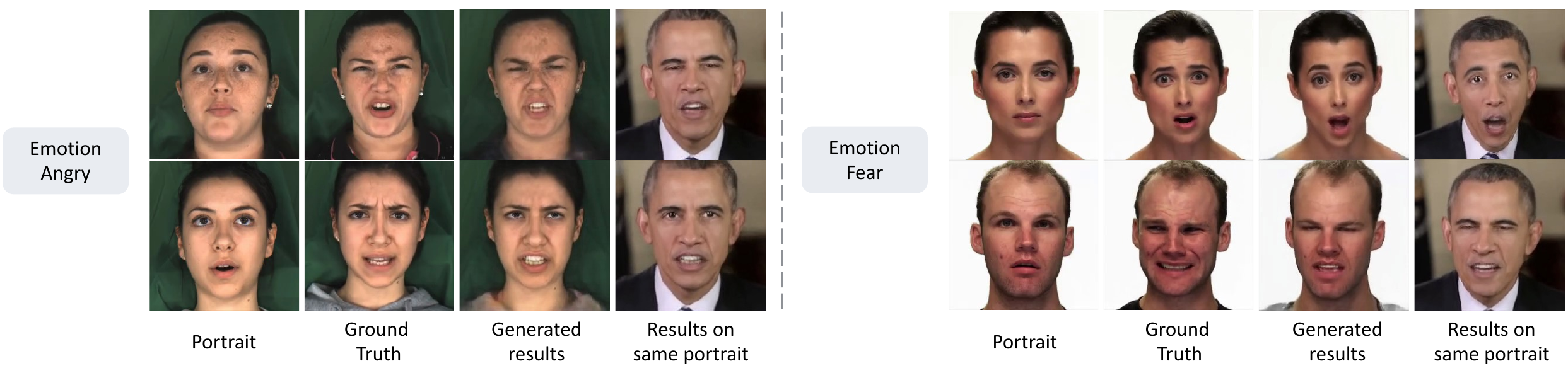}
  \caption{The results of speaking style prediction. The fourth column displays samples generated with predicted styles applied to the same portrait for clearer comparisons. 
  }
  \label{fig:style_predictor_qualitative_result}
\end{figure*}

\noindent\textbf{Evaluation of the Identity Preservation.} To evaluate the ability to preserve identity, we utilize a widely used metric CSIM. When computing the CSIM score, we utilize an off-the-shelf face recognition network ArcFace~\cite{deng2019arcface} to extract the deep identity features from each generated frame and then calculate the cosine similarities between the features of the input portrait and the generated frames. We find that when the ID in the style reference and portrait differ significantly, ID preservation worsens. So, we compute scores for StyleTalk and DreamTalk when the style reference comes from a randomly selected ID different from that in portrait. \cref{table:CSIM} shows the result on MEAD dataset. Wav2Lip attains the highest score since it merely changes the mouth region and leaves other parts intact. Non-emotional methods achieve better scores than emotional ones because changing emotions may change the identity perceived by humans and ArcFace. DreamTalk's score is competitive to non-emotional methods and is the best in emotional methods.

\noindent\textbf{Generalization Capabilities.} 
Leveraging the strong power of diffusion models, \method shows strong generalization capabilities. As shown in \cref{fig:songs} and \suppvideo, \method can even generate reasonable results for songs in multiple languages, even though our training dataset includes only a few multilingual audio and no song audio.  \suppvideo shows that \method also generalizes well to speech in various languages and noisy audio.


\noindent\textbf{Results of Speaking Style Prediction.} \cref{fig:style_predictor_qualitative_result} presents the results of speaking style predictions. The style predictor, utilizing emotional audio and neutral portraits, adeptly deduces personalized speaking styles aligned with those in the original videos. It can discern subtle expressions within the same emotion. For instance, for samples with angry emotion, the first-row speaker exhibits narrowed eyes, in contrast to the second-row speaker's intense, glaring stare. For samples with fear emotion, the first-row speaker's eyes and mouth are open, whereas the second-row speaker combines narrowed eyes with a contorted facial expression.

\begin{figure}[t!]
  \centering
  \includegraphics[width=0.47\textwidth]{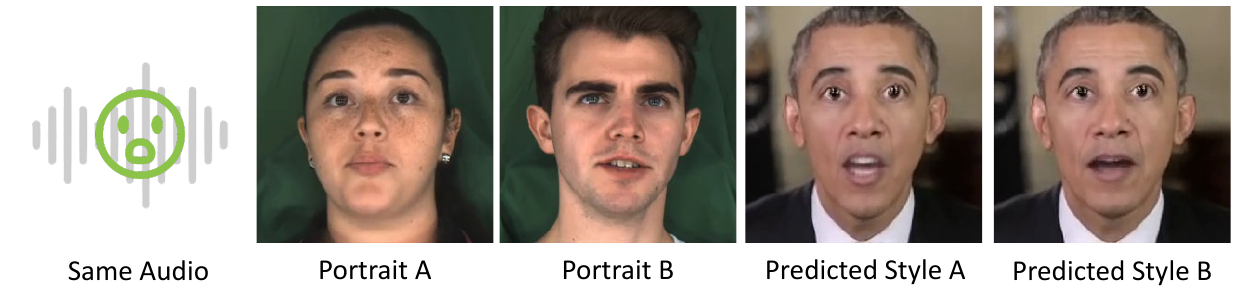}
  \caption{Analyzing the influence of portraits on style prediction. The audio conveys surprised emotion.}
  \label{fig:influence_of_portraits}
\end{figure}

We analyze the influence of portraits on speaking style prediction by predicting speaking styles with an audio clip and different input portraits. The predicted styles are subsequently applied to an identical portrait for a clearer comparison. As shown in \cref{fig:influence_of_portraits}, the predicted speaking styles match the subtle identity characteristics, such as gender, of the input portraits. The predicted style A generated more feminine results. This validates the necessity of integrating portrait information during style prediction.

\begin{figure*}[t]
    \centering\small
    \begin{minipage}[t!]{0.48\textwidth}
        \centering\small
        \includegraphics[width=\columnwidth]{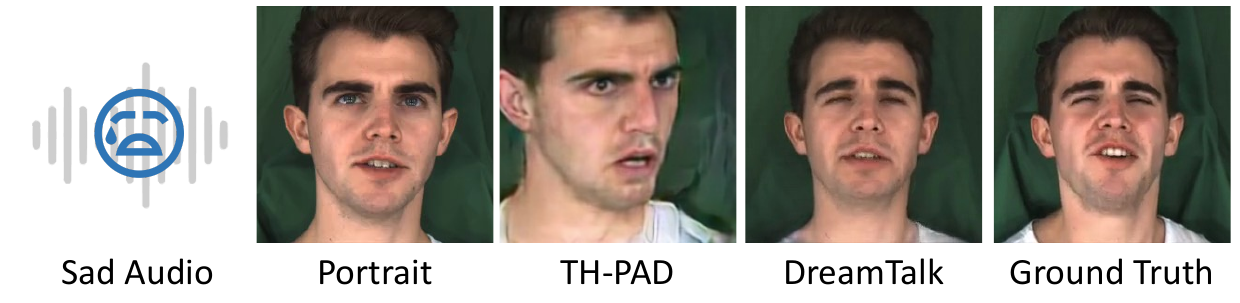}
        \caption{%
        Comparisons with TH-PAD.
        }
        \label{fig:compare_with_thpad}
    \end{minipage}
    \hskip1em
    \begin{minipage}[t!]{0.48\textwidth}
        \centering\small
        \includegraphics[width=\columnwidth]{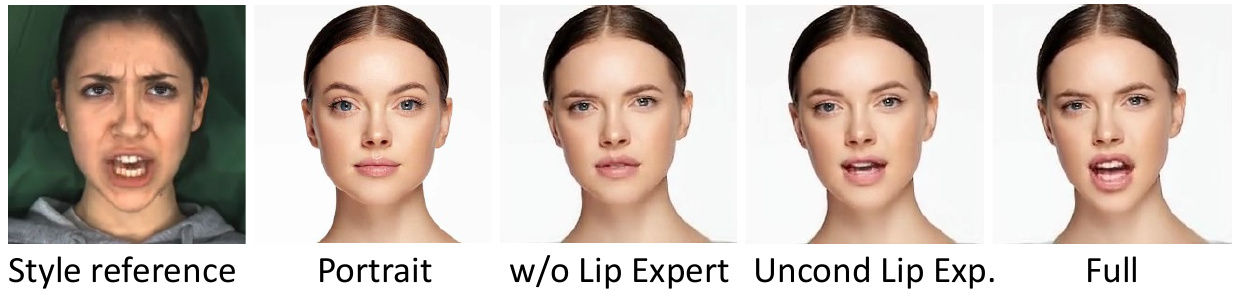}
        \caption{%
        Ablation study results of emotional talking head generation.
        }
        \label{fig:ablation}
    \end{minipage}
\end{figure*}



We compare the style predictor with TH-PAD~\cite{yu2023talking}, a method that also uses audio to predict expressions. We obtain TH-PAD samples with the authors' help. For comparisons, we use audio to predict speaking styles and use predicted styles to generate samples. We observe that TH-PAD fails to generate emotions conveyed in audio. TH-PAD only generates neutral expressions aligned with the audio rhythm.
\cref{fig:compare_with_thpad} shows that TH-PAD fails to generate the sad emotion reflected in the audio. We also conduct a quantitative comparison on MEAD. As shown in \cref{table:compare_with_thpad}, TH-PAD's SA and LMD scores, which measure emotion alignment, are inferior. 

\begin{table}[t]
        \centering\small
        \caption{Comparisons with TH-PAD on MEAD. DreamTalk(A) uses styles predicted from audio.
        }
        \setlength{\tabcolsep}{21pt}{
        \renewcommand{\arraystretch}{1.1}
\begin{tabular}{c|cc}
\toprule  
Method & F/M-LMD$\downarrow$ & SA$\uparrow$ \\
\shline
TH-PAD   & 6.38/5.81 & 5.3  \\
DreamTalk(A)  & \textbf{2.24}/\textbf{3.43} & \textbf{78.6} \\
\textcolor{gray}{Ground Truth} &  \textcolor{gray}{0 / 0} & \textcolor{gray}{92.5}  \\ 

\bottomrule 
\end{tabular}}
\label{table:compare_with_thpad}
\end{table}

\begin{table}[t]
        \centering\small
        \caption{The results of DreamTalk's ablation study on MEAD. CPBD is omitted due to no significant differences.}
        \setlength{\tabcolsep}{4pt}{
        \renewcommand{\arraystretch}{1.1}
\begin{tabular}{c|cccc}
\toprule  
Method & SSIM$\uparrow$ & F-LMD$\downarrow$ & M-LMD$\downarrow$  & $\text{Sync}_{\text{conf}}$$\uparrow$ \\
\shline
w/o Lip Expert   & 0.85 & \textbf{1.90} & 3.07   &  2.63 \\
Uncond Lip Expert  & 0.83 & 2.19 & 3.42    & \textbf{4.51}\\
\textbf{Full} & \textbf{0.86}  &  1.93 & \textbf{2.91}   & 3.78 \\

\bottomrule 
\end{tabular}}
\label{table:ablation_study}
\end{table}


\subsection{Ablation Study}
\label{sec:ablation_study}

\noindent\textbf{Emotional Talking Head Generation.} To analyze the contributions of our designs, we conduct an ablation study with two variants: (1) remove the style-aware lip expert (\textbf{w/o lip expert}); (2) trained with unconditional lip expert (\textbf{uncond lip expert}). Our full model is denoted as \textbf{Full}.    


\cref{fig:ablation} and \cref{table:ablation_study} present our ablation study results. The variant \textbf{w/o lip expert} exhibits a decline in lip-sync accuracy on the emotional dataset MEAD, despite its competitive F-LMD score indicating expressive facial generation. Conversely, \textbf{uncond lip expert} secures a superior SyncNet confidence score at the expense of speaking style expressiveness. The \textbf{Full} model achieves a harmonious balance, ensuring both precise lip synchronization and vivid expressions, thanks to the style-aware lip expert directing the diffusion model's expressive potential.


\begin{figure}[t!]
  \centering
  \includegraphics[width=0.47\textwidth]{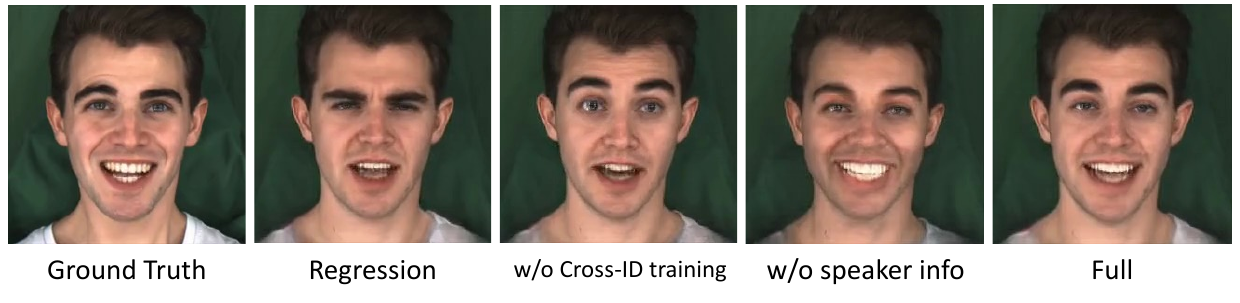}
  \caption{The qualitative results of style predictor's ablation study.}
  \label{fig:style_predictor_ablation}
\end{figure}


\noindent\textbf{Speaking Style Prediction.} To evaluate the impact of our design choices, we conduct an ablation study with three variants: (1) omitting speaker information and relying solely on audio for prediction (\textbf{w/o speaker info}); (2) during model training, the speaker info and audio are both obtained from the same video (\textbf{w/o cross-ID training}); (3) employing a regression model instead of a diffusion model for prediction (\textbf{regression}). Our full model is denoted as \textbf{Full}. When generating samples for evaluation, the facial images and audio we use are sourced from videos of the same individual expressing different emotions(\textit{e. g.} the face image is from a happy video while the audio is from an angry one.). This generation approach better aligns with real-world applications.   


 How to quantitatively evaluate the performance of speaking style prediction has not been explored before. we devise three metrics:
\begin{itemize}
    \item \textbf{Style Code Distance (SCD)} We extract the style codes from the videos that provide the audio input and compute the L2 distance between the predicted style codes and these style codes.
    \item \textbf{Motion Distance (MD)} We use the predicted style codes and the audio used for prediction to generate face motions and compute the L2 distance between the generated face motions and the face motions extracted from the ground truth videos. 
    \item \textbf{Style Accuracy (SA)} We use the SA metric mentioned in the \cref{sec:experimental_setup}. SA is evaluated on 3DMM face motions. The ground truth testing set gets $92.5\%$ accuracy.
\end{itemize}
We refrain from devising image-level metrics, such as training an image classifier for speaking style classification, due to several critical considerations. Firstly, factors in images that are irrelevant to expression, such as the speaker's identity and background elements, can adversely impact the accurate prediction of nuanced speaking styles. Secondly, inaccuracies introduced by the rendering process may further additionally hinder the accurate discernment of these subtle speaking styles.

\begin{table}[t]
\centering
\caption{The ablation study results of the style predictor.}
\setlength{\tabcolsep}{12pt}{
\renewcommand{\arraystretch}{1.1}
\begin{tabular}{c|ccc}
\toprule  
Method & SCD$\downarrow$ & MD$\downarrow$ & SA$\uparrow$  \\
\shline
w/o speaker info   & 0.49 & 0.28 & 64.3  \\
w/o cross-ID training  & 0.68 & 0.45 & 28.1  \\
regression  & 0.56 & 0.32 & 55.1 \\
\textbf{Full} & \textbf{0.42}  &  \textbf{0.23} & \textbf{78.6} \\

\bottomrule 
\end{tabular}}

\label{table:style_predictor_ablation_study}
\end{table}


The results are shown in \cref{table:style_predictor_ablation_study} and \cref{fig:style_predictor_ablation}.
The \textbf{w/o speaker info} variant successfully predicts emotions from audio but occasionally fails to maintain consistency between the predicted speaking style and speaker identity, leading to poor identity preservation. This underscores the importance of speaker information in predicting speaking styles. Although in experiments, we observed that \textbf{w/o cross-ID training} achieves slightly better performance than \textbf{Full} when the input portrait and audio are from the same video, it underperformed, often failing to predict the correct emotion, when inputs were from different videos. This suggests that identity 3DMM parameters may convey some expression information, and without cross-ID training, the model might derive emotional cues from this leaked information rather than the audio. The \textbf{regression} variant struggles to generate accurate expressions for certain data, highlighting the superior distribution-learning capability of diffusion models in facilitating speaking style prediction.

\subsection{Style Code Visualization}
\label{sec:style_code_visualization}
\begin{figure*}[t!]
  \centering
  \includegraphics[width=\textwidth]{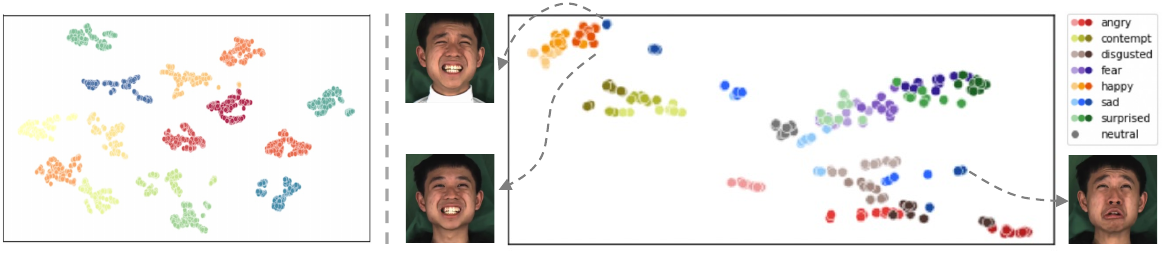}
  \caption{t-SNE visualization of style codes. \emph{Left}: Style codes from 15 speakers. Each color indicates style codes from an identical speaker. \emph{Right}: Style codes from a speaker, with darker hues representing increased emotional intensity.}
  \label{fig:combine_style_code_visualization}
\end{figure*}

\begin{figure}[t!]
  \centering
  \includegraphics[width=0.47\textwidth]{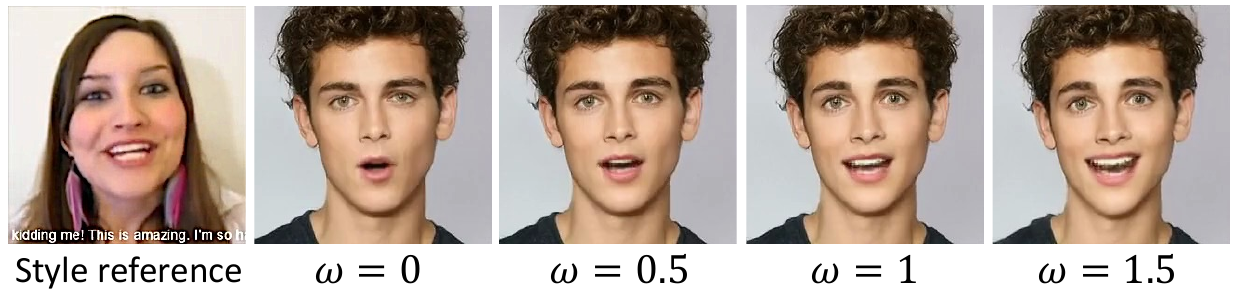}
  \caption{Modulating style intensity by adjusting the scale $\omega$ of classifier-free guidance.}
  \label{fig:cfg}
\end{figure}

We use t-SNE~\cite{van2008visualizing} to map style codes from the MEAD dataset's 15 speakers into a 2D space. Each speaker exhibits 22 distinct speaking styles, comprising seven emotions at three intensity levels, alongside a neutral style. For each style of each speaker, we randomly select 10 videos to extract style codes for visualization.

The left in \cref{fig:combine_style_code_visualization} shows that style codes first cluster based on identity rather than emotion, indicating the correlation between speaker identity and speaking styles, hence justifying the rationale for using portrait information to infer speaking styles. 

The right in \cref{fig:combine_style_code_visualization} shows that even within the same emotion and the same intensity, a speaker's expressions can vary significantly. The speaker expresses intense sadness in two ways. In one way, the speaker clenches teeth (top left portrait), similar to happy expressions (bottom left portrait), while in another way, the speaker depresses lip corners (bottom right portrait). The observation suggests  StyleTalk's assumption that speaking styles are consistent in videos with the same emotions is incorrect, which may impair performance.

\subsection{Modulating Style Intensity}

As elaborated in \cref{sec:dreamtalk_training}, the scale factor $\omega$ in the classifier-free guidance scheme can modulate the intensity of any input style, even those unseen during training. As shown in \cref{fig:cfg}, the style reference is in-the-wild and adjusting 
$\omega$ either amplifies or attenuates the designated style. When $\omega = 0$, \method produces a talking head with a neutral expression. We observed a noticeable decline in lip-sync accuracy when the scale factor $\omega$ exceeds $2$.


\subsection{User Study}

\begin{table}[t]
    \centering
    \caption{Mean ratings of user study with $95\%$ confidence intervals.}
    \setlength{\tabcolsep}{5pt}{
    \renewcommand{\arraystretch}{1.1}
    \begin{tabular}{c|ccc}
    \toprule
    Methods & Lip Sync$\uparrow$ & Realness$\uparrow$ & Style Consistency$\uparrow$ \\ 
        \shline
        MakeItTalk~\cite{zhou2020makelttalk} &  	2.73 $\pm$ 0.08 & 3.10 $\pm$ 0.08 & -  \\ 
        Wav2Lip~\cite{prajwal2020lip} &   			3.46 $\pm$ 0.07 & 2.45 $\pm$ 0.05 & - \\ 
        PC-AVS~\cite{zhou2021pose} & 				3.29 $\pm$ 0.04 & 2.86 $\pm$ 0.07 & -  \\ 
        AVCT~\cite{wang2022one} &  					3.43 $\pm$ 0.07 & 3.39 $\pm$ 0.08 & -  \\ 
        GC-AVT~\cite{liang2022expressive} & 		3.40 $\pm$ 0.06 & 2.21 $\pm$ 0.05 & 2.70 $\pm$ 0.05 \\ 
        EAMM~\cite{ji2022eamm} &  					2.83 $\pm$ 0.08 & 2.46 $\pm$ 0.07 & 2.43 $\pm$ 0.06 \\ 
        StyleTalk~\cite{ma2023styletalk} &  		3.42 $\pm$ 0.05 & 3.48 $\pm$ 0.07 & 2.88 $\pm$ 0.07  \\ 
        SadTalker~\cite{zhang2023sadtalker} &  	    3.44 $\pm$ 0.07 & 3.51 $\pm$ 0.06 & - \\ 
        PD-FGC~\cite{wang2023progressive} &  		3.16 $\pm$ 0.06 & 2.80 $\pm$ 0.06 & 3.06 $\pm$ 0.07 \\ 
        EAT~\cite{gan2023efficient} &   			3.39 $\pm$ 0.07 & 2.87 $\pm$ 0.08 & 2.91 $\pm$ 0.07 \\ 
        \hline
        \textbf{DreamTalk} &  						\textbf{3.80} $\pm$ 0.07 & \textbf{3.77} $\pm$ 0.07 & \textbf{3.34} $\pm$ 0.08  \\ 
        \textcolor{gray}{Ground Truth} &        \textcolor{gray}{4.21 $\pm$ 0.08} & \textcolor{gray}{4.09 $\pm$ 0.08} & - \\ 

    \bottomrule
    \end{tabular}}

    \label{tab:user_study}
\end{table}

\noindent\textbf{Emotional Talking Head Generation.} We conduct a user study to further evaluate our method. We generate 30 test samples for each method, which cover diverse speaking styles and speakers, and recruit 22 participants to rate samples.
Each participant is required to rate all samples (from 1 to 5, 5 is the best) on three aspects: (1) lip sync quality, (2) video realness, and (3) style consistency between the generated videos and the style reference (This metric is only evaluated on emotional methods. Since Ground Truth videos do not express the expressions reflected in the style references. The score for Ground Truth is omitted.). As shown in \cref{tab:user_study}, our method outperforms baselines across all aspects. A one-way ANOVA and a post-hoc Tukey test identify a significant difference ($p < 0.001$) between our method and other baselines on all aspects.

\noindent\textbf{Speaking Style Prediction.} In our user study, we evaluate the alignment between the original and predicted speaking styles. 
Directly assessing the alignment of speaking styles can be somewhat ambiguous, so we employ a comparative approach for evaluation.
Specifically, we create a series of video triplets. Each triplet consisted of a test video from our dataset and two generated videos. 
The first video was generated using a style code predicted from an input portrait, sharing the same speaker identity as in the test video but displaying a neutral emotion, combined with the audio from the test video. The second video is generated using the style code extracted from videos with the same emotion but from a speaker different from the one in the test video. We recruit 20 participants. Each participant is then asked to evaluate 20 triplets and identify which of the generated videos most accurately reflected the speaking style of the test video. 
The videos generated using predicted style codes are preferred in $75.8\%$ of all ratings. This indicates that the style predictor is able to infer personalized speaking styles that are aligned with the audio. 


\section{Limitations}

Despite DreamTalk's promising advancements in emotional talking head generation, it encounters several challenges that open avenues for future research.

When using a constant style reference, DreamTalk generates expressions that are strictly consistent with the main expressions in the reference but lack expression changes over time, such as eye blinking. Generating emotions with rich temporal variations is, compared to methods that generate neutral or coarse-grained emotions, more difficult for methods that achieve fine-grained emotional control, like DreamTalk. This is because these methods must achieve both precise control of expressions and diverse changes. To address the issue, temporal changes in expressions can be achieved by using temporally changing style references when generating each video frame. As discussed in \supp, DreamTalk can achieve smooth expression changes by changing the style references.  Eye blinks can be achieved by using eye blink loss~\cite{zhang2023sadtalker} during training or post-editing 3DMM, a common practice used in previous methods~\cite{cudeiro2019capture,peng2023emotalk} that also aim to control expressions. Specifically, we can obtain the parameter changes of blinking and then edit the generated expression parameters. A video with the eye blinking through post-editing is shown in \suppvideo.

DreamTalk may change the speaker's identity when the identity in the reference video highly differs from that in the portrait. The reason is that 3DMM expression parameters leak identity information. This leakage leads to the generated identity becoming somewhat similar to the identity in the reference. The issue can be mitigated by adopting 3DMM which decouples expression and identity more effectively.

The style predictor sometimes struggles with accurately identifying emotions in low-emotion-intensity audio clips from the MEAD dataset. The reason is that in some MEAD videos, the audio does not correspond with the expressed emotions. To enhance prediction accuracy, it is beneficial to employ a dataset where the audio closely aligns with the expressed emotions. Another solution is to incorporate textual information from audio during prediction, a strategy commonly employed in speech emotion recognition~\cite{wang2023exploring}.

DreamTalk occasionally produces artifacts around the mouth area, such as teeth flickering, particularly during intense expressions. The issue comes from the renderer and can be mitigated by using more advanced renderers.

Despite these challenges, DreamTalk marks a significant stride in the realm of emotional talking head generation, paving the way for further innovations.


\section{Conclusion}
In this work, we propose \method, a novel diffusion-based framework that can consistently generate high-quality talking heads in diverse speaking styles and conveniently use audio to specify personalized emotions.
We develop a denoising network for creating emotional, audio-driven facial motions and introduce a style-aware lip expert to optimize lip-sync while maintaining emotion intensity. 
Additionally, we devise a style predictor that infers speaking styles directly from audio, eliminating the need for video references. 
Extensive experiments validate the efficacy of \method.
The results demonstrate that employing diffusion models markedly improves the quality of emotional talking head generation.
\section{Ethical Consideration}

\method can generate vivid talking head videos, opening up diverse applications but also posing risks like promoting hatred or depicting violence. Misuse could harm individuals or groups, perpetuate stereotypes, and spread misinformation. To mitigate these risks, we've implemented safeguards like watermarks on all outputs and advising against using images without consent. We remain committed to continuous research to minimize adverse societal impacts.



\bibliographystyle{IEEEtran}
\bibliography{main}

\end{document}